\documentclass[11pt]{article}
\usepackage[T1]{fontenc}
\usepackage[utf8]{inputenc}
\usepackage{lmodern}
\usepackage[a4paper,margin=1in]{geometry}
\usepackage{amsmath,amssymb,amsthm,mathtools}
\usepackage{enumitem}
\usepackage[hidelinks]{hyperref}
\usepackage{xcolor}
\usepackage{graphicx}
\usepackage{booktabs}
\usepackage{float}
\usepackage{titlesec}

\titleformat{\section}
  {\normalfont\Large\bfseries}{\thesection}{1em}{}
\titleformat{\subsection}
  {\normalfont\large\bfseries}{\thesubsection}{1em}{}

\hypersetup{pdftitle={Diagnostic Benchmarks for Invariant Learning Dynamics}, pdfauthor={Datorien Laurae Anderson}}

\title{
    \textbf{Diagnostic Benchmarks for Invariant Learning Dynamics} \\
    \large Empirical Validation of the Eidos Architecture
}
\author{Datorien Laurae Anderson \\ Occybyte \\ \href{mailto:datorien@occybyte.com}{datorien@occybyte.com}}
\date{December 22, 2025 \\ (Updated February 8, 2026)}

\theoremstyle{definition}

\begin{document}
\maketitle

\begin{abstract}
We present the PolyShapes-Ideal (PSI) dataset, a suite of diagnostic benchmarks designed to isolate topological invariance---the ability to maintain structural identity across affine transformations---from the textural correlations that dominate standard vision benchmarks. Through three diagnostic probes (polygon classification under noise, zero-shot font transfer from MNIST, and geometric collapse mapping under progressive deformation), we demonstrate that the Eidos architecture achieves $>$99\% accuracy on PSI and 81.67\% zero-shot transfer across 30 unseen typefaces without pre-training. These results validate the ``Form-First'' hypothesis: generalization in structurally constrained architectures is a property of geometric integrity, not statistical scale.
\end{abstract}

\section{Motivation: Isolating Geometric Invariants}

Standard machine learning benchmarks, while valuable for comparing aggregate performance, often fail to distinguish between \textit{statistical correlation} (texture, local features) and \textit{invariant reasoning} (topology, global structure) \cite{geirhos2019texture}. A model may achieve high accuracy on ImageNet by creating ``texture atlases'' rather than understanding object permanence or shape, leading to generalization failures on new test sets \cite{recht2019imagenet}. 

Standard convolutional networks are biased toward texture over shape \cite{geirhos2019texture}. The \textbf{Form-First} hypothesis \cite{anderson2026invariant} inverts this default: learning should prioritize the discovery of topological invariants over instance-specific details. To rigorously validate this hypothesis, we constructed a suite of diagnostic probes. These tasks are designed to be ``primitive'' yet structurally demanding, stripping away the textural correlations that allow texture-biased networks to bypass global reasoning. 

\paragraph{Scope.} The Eidos architecture \cite{anderson2026invariant} belongs to a class of discrete-commitment systems that select ternary states rather than interpolating continuous activations. Terms used throughout this paper---\textit{Structural Tension}, \textit{Geometric Collapse}, \textit{basin of attraction}---are drawn from dynamical systems theory \cite{strogatz2015nonlinear} and adapted for this architecture's internal states, not used as metaphors for loss gradients. Where our terminology introduces new distinctions, we provide explicit connections to established concepts in Section~\ref{sec:glossary}.

The \textbf{PolyShapes-Ideal (PSI)} dataset, for instance, removes natural image biases to isolate the fundamental problem of boundary closure and vertex counting. Success on this benchmark requires the model to organize information into a cohesive global representation, as no local shortcut exists to distinguish a heptagon from an octagon in the presence of noise. This allows us to observe the emergence of generalization not as a gradual statistical improvement, but as a distinct phase transition in the system's internal dynamics.

\section{PolyShapes-Ideal (PSI): Inductive Bias Alignment}

The development of the PSI benchmark reveals the interaction between data incentives and architectural priors. We observed three distinct learning regimes corresponding to the information structure of the training data.

\subsection{PSI v0.1: The Limits of Implicit Structure}
In the initial regime, the model was trained exclusively on ``filled'' polygons ($n=3$ to $10$) with heavy noise, while ``outline'' and ``ideal'' (regular) variants were held out. The hypothesis was that the model would infer the boundary as the causal generator of the fill.

Empirically, the model converged on a specific heuristic: ``mass implies class.'' It learned that filled regions defined the manifold, while treating outlines as out-of-distribution noise. Consequently, outline accuracy hovered at or below chance levels for the majority of training. True invariance emerged only via a late-stage \textbf{phase transition} (approx.\ epoch 8/10), where validation accuracy spiked to $\sim$70\%. Phase transitions in learning dynamics---sharp, non-linear jumps in generalization performance---have been documented in both deep linear networks \cite{saxe2014exact} and algorithmic datasets \cite{power2022grokking}. The transition observed here differs from grokking in timing: it occurs \emph{before} the model has memorized the training distribution, suggesting the structural reorganization documented in \cite{anderson2026invariant} rather than delayed generalization from overfitting. This delay indicates that without explicit boundary data, the model prioritizes lower-energy texture heuristics before reorganizing into a topological representation.

\subsection{PSI v0.2: Explicit Boundary Constraints}
PSI v0.2 introduced outline samples directly into the training distribution. This intervention explicitly invalidated the ``mass-only'' shortcut. The behavioral shift was immediate:
\begin{enumerate}
    \item \textbf{Rapid Convergence:} Outline generalization tracked training accuracy from the first epoch.
    \item \textbf{Elimination of Hesitation:} The ``hesitation phase'' observed in v0.1 vanished, replaced by a smooth co-evolution of training and zero-shot metrics.
    \item \textbf{Ideal Generalization:} Recognition of perfect ``Ideal'' polygons stabilized in the 15--25\% range.
\end{enumerate}
This confirms that the architecture possesses the requisite capacity for topological reasoning ($\sim$490k parameters), but the activation of this capacity is contingent on the falsification of simpler statistical explanations.

\subsubsection{Volumetric Extension (Polyhedra Panic)}

We extended the boundary discrimination task to $\mathbb{R}^3$ with \textbf{Polyhedra Panic (PSIP)}, challenging the model to infer volumetric classes (tetrahedron, octahedron, high-order geodesic spheres) from 2D projections. 

Contrary to the expectation that increased dimensionality would hinder learning, PSIP demonstrated \textbf{hyper-rapid convergence}. 
\begin{itemize}
    \item \textbf{Epoch 1:} Validation accuracy reached \textbf{92.3\%}, with ``Wireframe'' recognition at 84.4\% and ``Shaded'' recognition at 93.8\%.
    \item \textbf{Epoch 4:} Validation accuracy peaked at \textbf{97.64\%}, with Wireframe accuracy at \textbf{98.52\%}.
\end{itemize}

This suggests that the Eidos architecture's native representation is homeomorphic to volumetric structures. The transition from 2D to 3D reduced the structural tension \cite{anderson2026invariant} of the problem, making 3D reasoning a lower-energy state than 2D boundary discrimination. We observe a ``criticality state'' where high-$N$ polyhedra become indistinguishable from differentiable manifolds (spheres), mirroring the continuum limit in calculus.

\section{Zero-Shot Invariant Transfer: The Font Probe}

To verify that the model has learned the \textit{concept} of a digit rather than the \textit{distribution} of MNIST handwriting, we executed a zero-shot transfer experiment. A model trained to 99.36\% accuracy on MNIST was evaluated against a suite of 30 unseen system fonts. 

\subsection{Methodology}
The key distinction here is topological. Handwriting involves motor noise and stroke variability, whereas fonts involve stylized rigidities, serifs, and geometric deformations. If the model relies on pixel-template matching, transfer should be minimal (near chance). This serves as our primary falsification test against geometric priors. High transfer accuracy implies the abstraction of a \textbf{Platonic Form} \cite{anderson2026invariant}---a representation invariant to local affine transformations. Low transfer accuracy effectively falsifies the hypothesis, as it would indicate that the architecture failed to extract transferrable invariants.

The font selection was curated to span three distinct typological categories: handwriting-adjacent, stylized lettering, and monospace. To ensure robustness, we employed a staged evaluation protocol: initially testing on a 4-font subset to verify basic transfer, before expanding to a 25-font suite to test universality across diverse typographies.

The experimental protocol is as follows:
\begin{enumerate}
    \item \textbf{Baseline Verification:} Train the model on MNIST (handwriting) exclusively to verify architectural stability \cite{lecun1998gradient}.
    \item \textbf{Dataset Curation:} Utilize a suite of 25 distinct Google Fonts across the target categories.
    \item \textbf{Zero-Shot Dual-Path Training:} Execute the 'MNIST-to-Fonts' training script, evaluating zero-shot performance on the font suite at every epoch across multiple independent runs.
    \item \textbf{Data Compilation:} Aggregate performance metrics to identify stability regimes and transfer thresholds.
    \item \textbf{Checkpoint Probing:} Examine saved state vectors to map the decision boundaries.
\end{enumerate}

\subsection{Results}
The model achieved an aggregate zero-shot accuracy of \textbf{81.67\%} across 30 fonts.

\begin{figure}[H]
    \centering
    \begin{minipage}{0.48\textwidth}
        \centering
        \includegraphics[width=\linewidth]{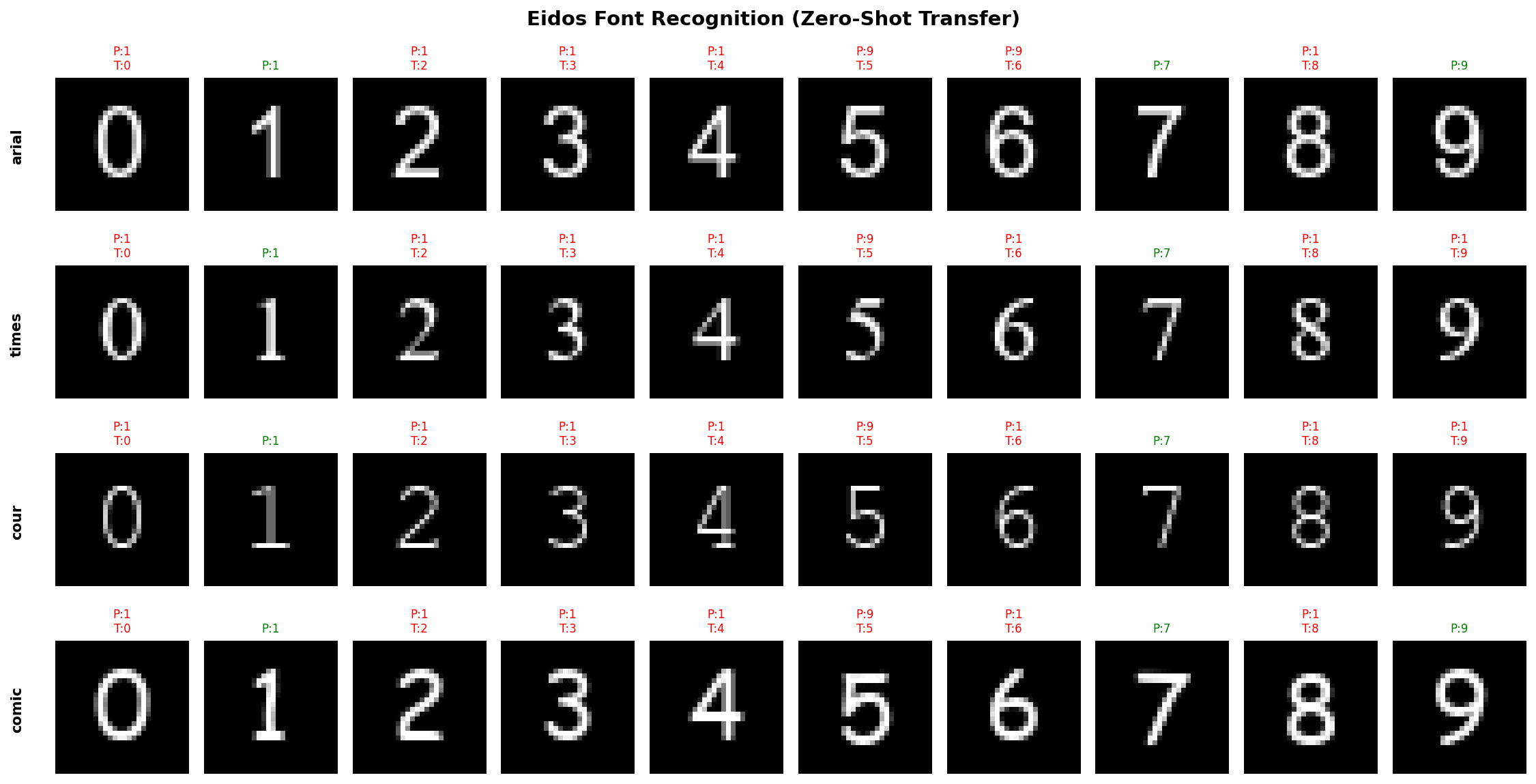}
        \caption{\textbf{Hybrid Architecture (v0.1).} Zero-shot font recognition using an early hybrid Eidos variant that retains classical statistical layers. Each cell renders a system font digit (rows: Arial, Times New Roman, Courier New, Comic Sans; columns: digits 0--9). Labels above each cell show \textcolor{green}{\textbf{P:$n$}} (predicted) and \textcolor{red}{\textbf{T:$n$}} (true class); green indicates correct classification, red indicates misclassification. The hybrid model collapses digits 5, 6, 7, and 9 into a narrow attractor basin around digit~1, demonstrating that residual classical layers destroy the topological discrimination required for transfer.}
        \label{fig:archbefore}
    \end{minipage}\hfill
    \begin{minipage}{0.48\textwidth}
        \centering
        \includegraphics[width=\linewidth]{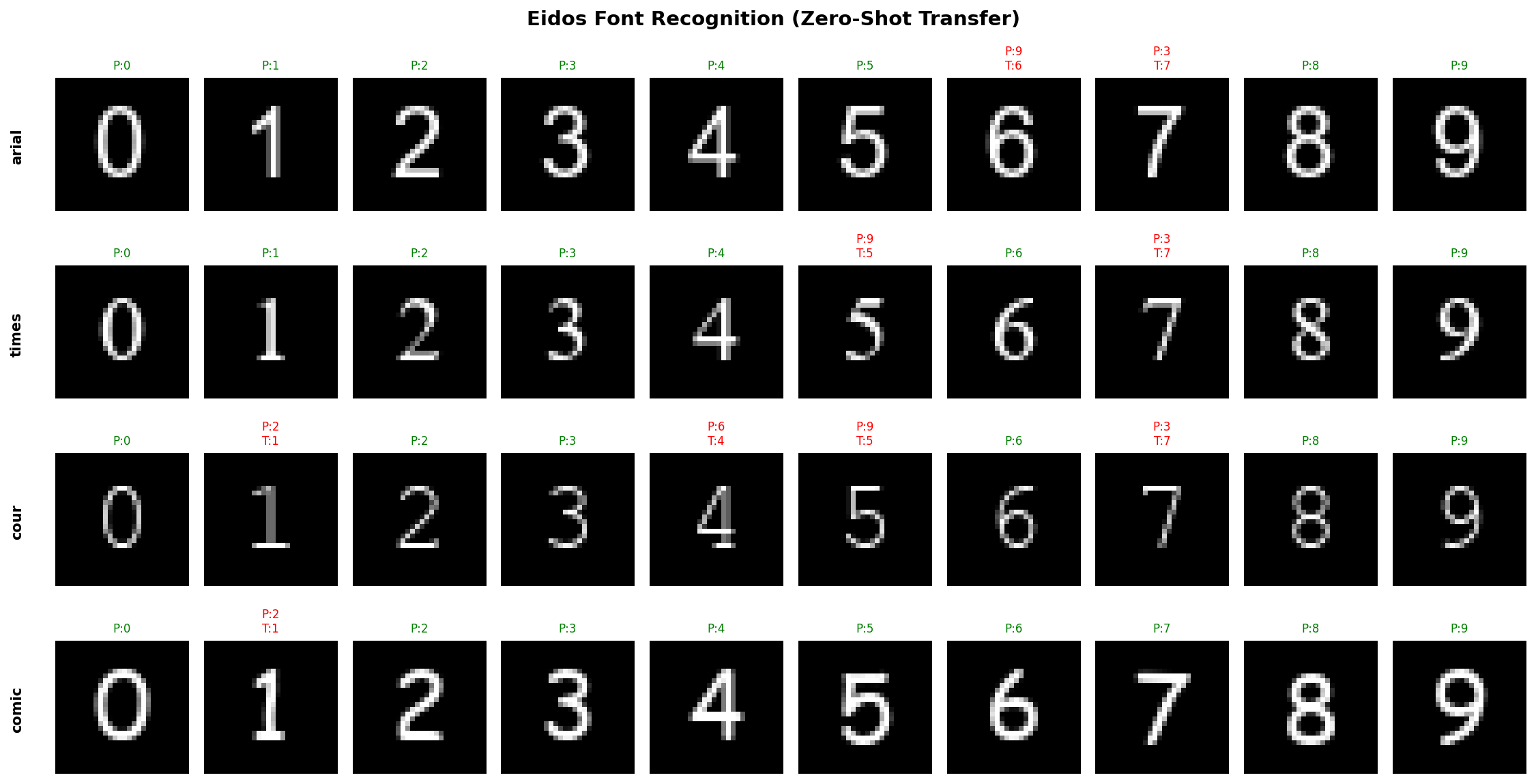}
        \caption{\textbf{Pure Eidos Architecture.} Same protocol as Figure~\ref{fig:archbefore}, evaluated after full architectural purification (removal of all classical interpolation layers). The model correctly identifies the majority of digits across all four typefaces. Residual errors concentrate on digits 6 and 7---topologically confusable forms whose open/closed loop distinction is stylistically ambiguous in certain fonts---while structurally unambiguous digits (0, 2, 3, 4, 8) achieve perfect transfer.}
        \label{fig:archafter}
    \end{minipage}
\end{figure}

\begin{table}[H]
\centering
\caption{\textbf{Zero-Shot Font Recognition Metrics.} The model demonstrates robust transfer to both serif and sans-serif families, with degradation only in highly stylized ``display'' fonts (e.g., Arial Narrow Bold).}
\label{tab:font30}
\resizebox{\textwidth}{!}{%
\begin{tabular}{lc|lc|lc}
\toprule
\textbf{Font Family} & \textbf{Accuracy} & \textbf{Font Family} & \textbf{Accuracy} & \textbf{Font Family} & \textbf{Accuracy} \\
\midrule
Arial (Std) & 70\% & Amiri (Serif) & 80-90\% & Arial Narrow & 60-80\% \\
Times New Roman & 90\% & Antiqua & 100\% & Algerian & 80\% \\
Courier New & 80\% & Comic Sans & 100\% & Arial Nova & 70-80\% \\
Agency FB & 70-80\% & Alef & 70-90\% & System Default & 80\% \\
\bottomrule
\end{tabular}
}
\end{table}

Notable findings:
\begin{itemize}
    \item \textbf{Perfect Transfer (100\%):} Observed in Antiqua and Comic Sans, suggesting their geometries align closely with the ``mean topological form'' of handwritten digits found within the MNIST dataset.
    \item \textbf{Robustness to Serifs:} High performance on Times New Roman (90\%) and Amiri (90\%) indicates the model ignores decorative serif artifacts.
    \item \textbf{Inverse Transfer:} Training on Fonts and testing on MNIST yields ${\sim}$64.89\% accuracy (vs.\ 99.38\% supervised). This asymmetry suggests that the handwritten domain covers a broader topological variance than the rigid font domain, effectively encompassing the ``font manifold,'' whereas fonts are a subset of possible forms.
\end{itemize}

\section{Geometric Collapse Manifolds}

Loss landscape geometry determines how stably a network occupies its learned solutions \cite{li2018visualizing}. In contrastive and self-supervised settings, \textit{dimensional collapse}---where representations converge to a low-rank subspace---degrades downstream performance \cite{jing2022understanding}. We extend this perspective to discrete-commitment architectures by mapping \textbf{Geometric Collapse Thresholds} \cite{anderson2026invariant}: the affine deformation magnitude at which the model's internal representation loses topological integrity.

By progressively applying affine shear and rotation to held-out digit images, we identify the precise boundary where confidence collapses. Two distinct classes of manifold stability emerge:
\begin{itemize}
    \item \textbf{Deep Attractors (Robust):} Digits with closed topological loops (0, 6, 8, 9) exhibit wide basins of attraction \cite{strogatz2015nonlinear}. They retain their identity under significant deformation.
    \item \textbf{Fragile Manifolds (Meta-Stable):} Open-form digits (1, 7) rely on specific angular relations. These manifolds are shallow; minor deformations push the representation across the separatrix into noise or neighbor classes.
\end{itemize}

This differential stability confirms that Eidos encodes \textbf{structural tension}—the preservation of loops and intersections—rather than pixel density.

\begin{figure}[H]
    \centering
    \includegraphics[width=0.9\textwidth]{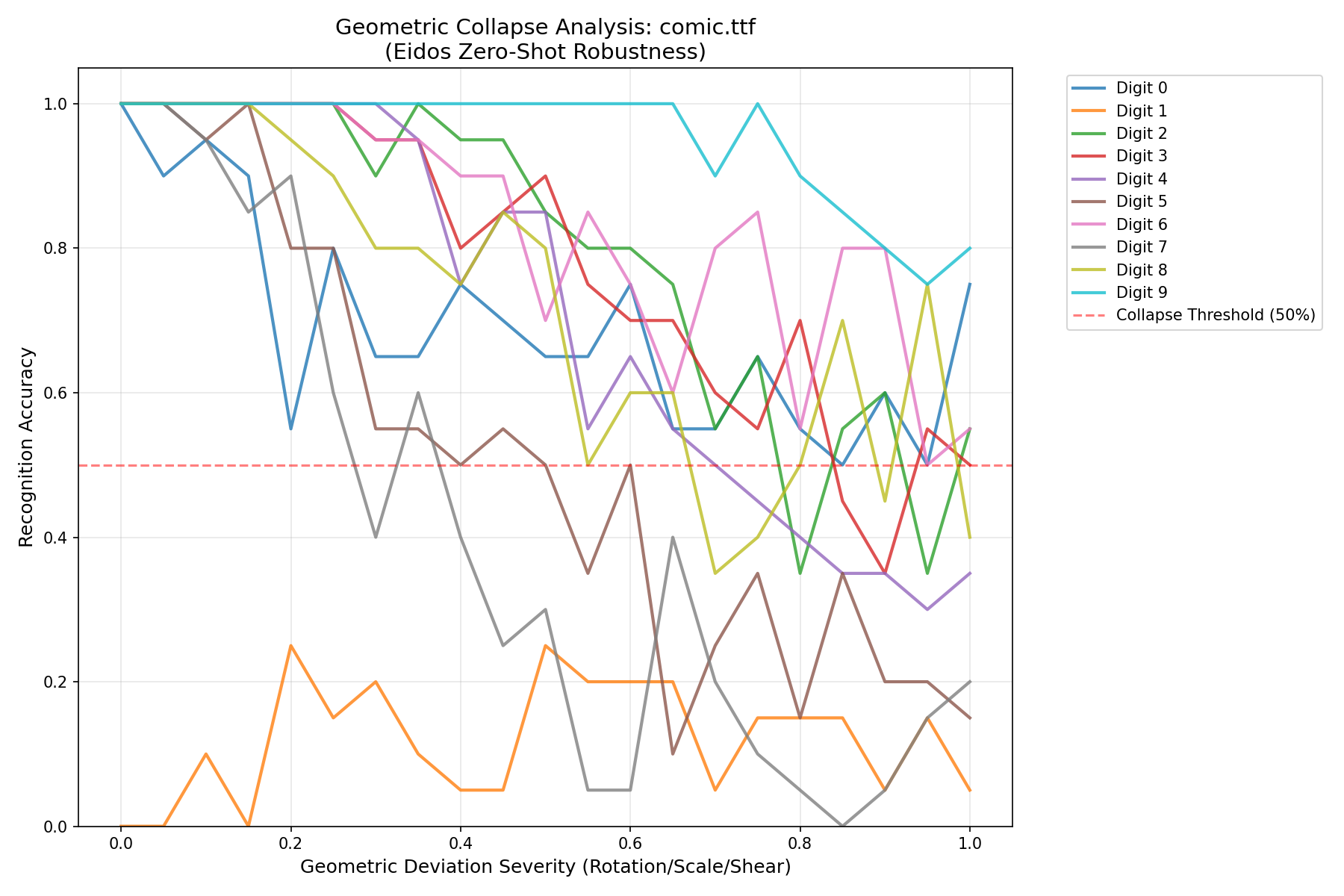}
    \caption{\textbf{Geometric Collapse Manifolds.} Recognition accuracy per digit class under progressive affine deformation (combined rotation, scale, and shear) applied to Comic Sans renderings. \textit{X-axis:} Geometric Deviation Severity, normalized from 0.0 (no deformation) to 1.0 (maximum). \textit{Y-axis:} Recognition accuracy (proportion of correct classifications at each severity level). The dashed red line marks the \textbf{Collapse Threshold} at 50\% accuracy---below this, the model's commitment to the correct class is no better than chance. \textbf{Deep Attractor} digits (0, 9: closed loops with high topological genus) remain above the collapse threshold across the full severity range, demonstrating wide basins of attraction. \textbf{Fragile} digits (1, 7: open strokes with no closed loops) collapse below threshold before severity reaches 0.2, confirming that their invariant representations depend on angular relations that are destroyed by small shear perturbations. Intermediate digits (2, 3, 4, 5, 6, 8) show graded degradation proportional to their topological complexity---digits with partial closures (6, 8) degrade more slowly than those relying on curvature alone (3, 5).}
    \label{fig:stress_collapse}
\end{figure}

\section{Conclusion}

The Eidos architecture demonstrates that \textbf{invariant learning} is distinct from statistical approximation. Through the diagnostic lenses of PSI, Zero-Shot Font Transfer, and Geometric Collapse Mapping, we have shown:
\begin{enumerate}
    \item \textbf{Form Emergence:} Invariants emerge via early phase transitions only when simple statistical shortcuts are structurally invalidated (PSI v0.2). Without explicit boundary data, the model defaults to texture heuristics; with it, topological reasoning activates immediately.
    \item \textbf{Geometric Generalization:} Learned representations transfer zero-shot to topologically similar but texturally distinct domains (30 unseen typefaces at 81.67\% aggregate accuracy), validating the extraction of transformation-invariant forms.
    \item \textbf{Differential Stability:} Stress-testing under progressive affine deformation reveals that digits with closed topological loops (0, 6, 8, 9) exhibit deep attractor basins, while open-form digits (1, 7) collapse rapidly---confirming that the architecture encodes structural properties rather than pixel distributions.
\end{enumerate}

These findings argue for a \textbf{Diagnostic-First} approach to architecture evaluation, where models are assessed not only on leaderboard accuracy but on the topological validity of their internal representations.

\section{Semantic Disambiguation}
\label{sec:glossary}
To ensure clarity for researchers accustomed to statistical learning paradigms, we define the specific geometric-dynamic terms used in this paper and relate them to established concepts.

\begin{description}
    \item[Form] \hfill \\
    \textit{Standard ML:} A collection of features or a class label. \\
    \textit{Eidos} \cite{anderson2026invariant}: A topologically closed, self-consistent boundary graph (e.g., a ``triangle'' is the closure of 3 vertices, not a texture). Form is the invariant structure that persists across affine transformations.
    
    \item[Structural Tension] \hfill \\
    \textit{Standard ML:} Error or Loss (scalar). \\
    \textit{Eidos:} The energetic cost required to maintain a deformation, analogous to the curvature of the loss landscape around a minimum \cite{li2018visualizing}. A ``Square'' deformed into a ``Trapezoid'' holds high tension; if the deformation exceeds the basin of attraction, the tension breaks, and the Form collapses.
    
    \item[Geometric Collapse] \hfill \\
    \textit{Standard ML:} Misclassification, or dimensional collapse in representation learning \cite{jing2022understanding}. \\
    \textit{Eidos:} A phase transition where the internal representation loses its topological integrity. It is not a smooth degradation of probability but a discrete failure of the attractor to hold the state.
    
    \item[Invariant] \hfill \\
    \textit{Standard ML:} A feature robust to noise (e.g., edge detection). \\
    \textit{Eidos:} A property that remains constant under a defined group of transformations (Rotation, Scale, Projection). Invariants are reasoned, not approximated.
\end{description}

\section*{Supplementary Material}
This work is supported by open-source reproducibility artifacts, including training logs, model checkpoints, and the experimental scripts used to generate these results \cite{anderson2026invariant}.

\bibliographystyle{plain}
\bibliography{references}

\end{document}